\documentclass[letterpaper, 10 pt, conference]{ieeeconf}  

\IEEEoverridecommandlockouts                              

\usepackage[bookmarks=true]{hyperref}
\usepackage{amsmath,amssymb,amsfonts}
\usepackage{algorithmic}
\usepackage{graphicx}
\usepackage{textcomp}
\usepackage{xcolor}
\usepackage{multicol}
\usepackage[font=small]{caption}
\usepackage{color}
\usepackage{colortbl}
\usepackage{subcaption}
\usepackage{makecell}
\usepackage[bookmarks=true]{hyperref}
\usepackage{bbm}
\usepackage{comment}

\definecolor{LightCyan}{rgb}{0.776470,0.843137,0.933333}
\definecolor{light-gray}{rgb}{0.9,0.9,0.9}
\definecolor{jb_green}{rgb}{0.772549, 0.878431, 0.705882}
\title{\LARGE \bf
Snake Robot with Tactile Perception \\ Navigates on Large-scale Challenging Terrain
}

\author{Shuo Jiang$^{}$, Adarsh Salagame$^{}$, Alireza Ramezani$^{}$, and Lawson L.S. Wong$^{}$
\thanks{$^{}$ The authors are with Northeastern University, Boston, MA, USA
        {\tt\small $\{$ shuo.jiang, salagame.a, a.ramezani, l.wong$\}$@northeastern.edu}}%
}

\begin{document}

\maketitle
\thispagestyle{empty}
\pagestyle{empty}

\begin{abstract}

Along with the advancement of robot skin technology, there has been notable progress in the development of snake robots featuring body-surface tactile perception. In this study, we proposed a locomotion control framework for snake robots that integrates tactile perception to augment their adaptability to various terrains. Our approach embraces a hierarchical reinforcement learning (HRL) architecture, wherein the high-level orchestrates global navigation strategies while the low-level uses curriculum learning for local navigation maneuvers. Due to the significant computational demands of collision detection in whole-body tactile sensing, the efficiency of the simulator is severely compromised. Thus a distributed training pattern to mitigate the efficiency reduction was adopted. We evaluated the navigation performance of the snake robot in complex large-scale cave exploration with challenging terrains to exhibit improvements in motion efficiency, evidencing the efficacy of tactile perception in terrain-adaptive locomotion of snake robots.

\end{abstract}

\section{INTRODUCTION}

With the recent advancements in the field of bionic robots, snake robots have drawn increasing attention. These robots emulate the body structure of snakes, comprising sequentially interconnected joints. This unique body configuration affords them the capability to execute distinctive motions. Furthermore, owing to their slender body shape, they excel in accessing narrow environments that prove challenging for other types of robots. Consequently, snake robots have demonstrated commendable performance in a range of specialized scenarios, including underwater environments \cite{kelasidi2016innovation}, earthquake rescue \cite{whitman2018snake}, etc.

Snake robots primarily rely on undulating motions to generate anisotropic friction on the contact surface for propulsion. In contrast to legged or wheeled robots, this mode of motion yields a plethora of ground contacts that pervade the entire body, presenting challenges in dynamics analysis. However, these intricate ground contacts serve as informative conduits for capturing various terrain characteristics, including surface roughness and slope. Early control strategies \cite{sihite_unsteady_2022,sihite_unilateral_2021,ramezani_generative_2021,lessieur_mechanical_2021,de_oliveira_thruster-assisted_2020,grizzle_progress_nodate,sihite_multi-modal_2023,sihite_orientation_2021,ramezani_towards_2020} for shape-shifting robots employed dynamic modeling and feedback control, often leveraging serpenoid curves \cite{rezapour2014path} or backbone curves \cite{elsayed2021mobile} to design gait patterns. These approaches allowed for the generation of diverse movements, including sidewinding, undulation, or lateral rolling. It is important to note that these control methodologies were predominantly validated for path tracking on flat terrain. Among these strategies, the Central Pattern Generator (CPG) \cite{bing2017cpg} emerged as a comprehensive approach for gait generation and transition. By outputting sinusoidal curves with distinct parameters, CPG could generate a variety of snake robot gaits. Subsequently, the concept of segmented control was introduced \cite{sanfilippo2016virtual}. The body of the snake robot can be divided into multiple segments according to its functionality. Employing different gaits in each segment unveiled the potential for the discovery of more complex gaits, such as the C-pedal wave and crawler gait \cite{takemori2021hoop}.

\begin{figure}[t]
\centering
\includegraphics[width=\linewidth]{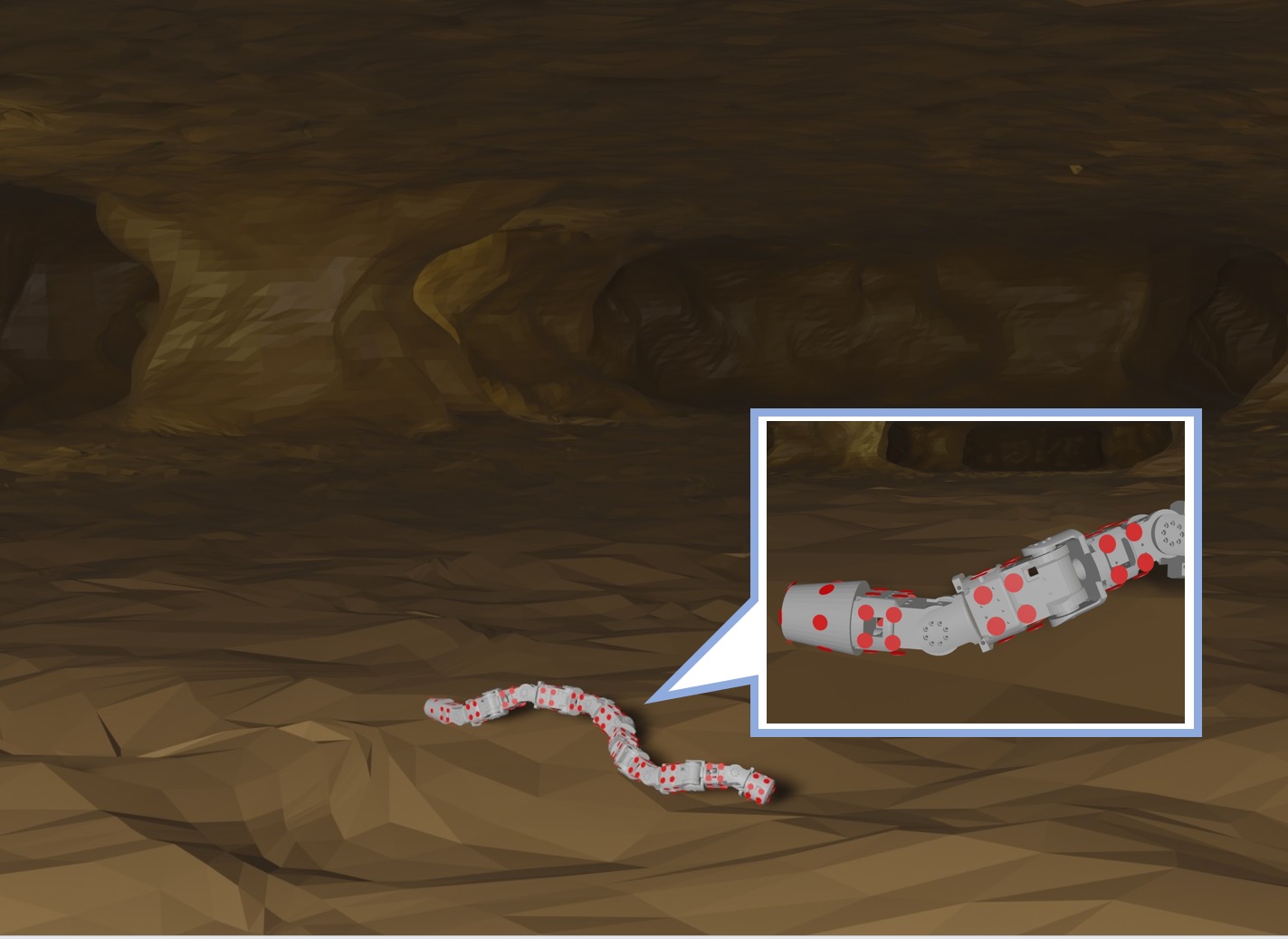}
\caption{Snake robot with whole-body tactile sensors performing cave navigation.}
\label{fig_1}
\vspace{-10pt}
\end{figure}

\begin{figure*}[t]
\centering
    \includegraphics[width=\linewidth]{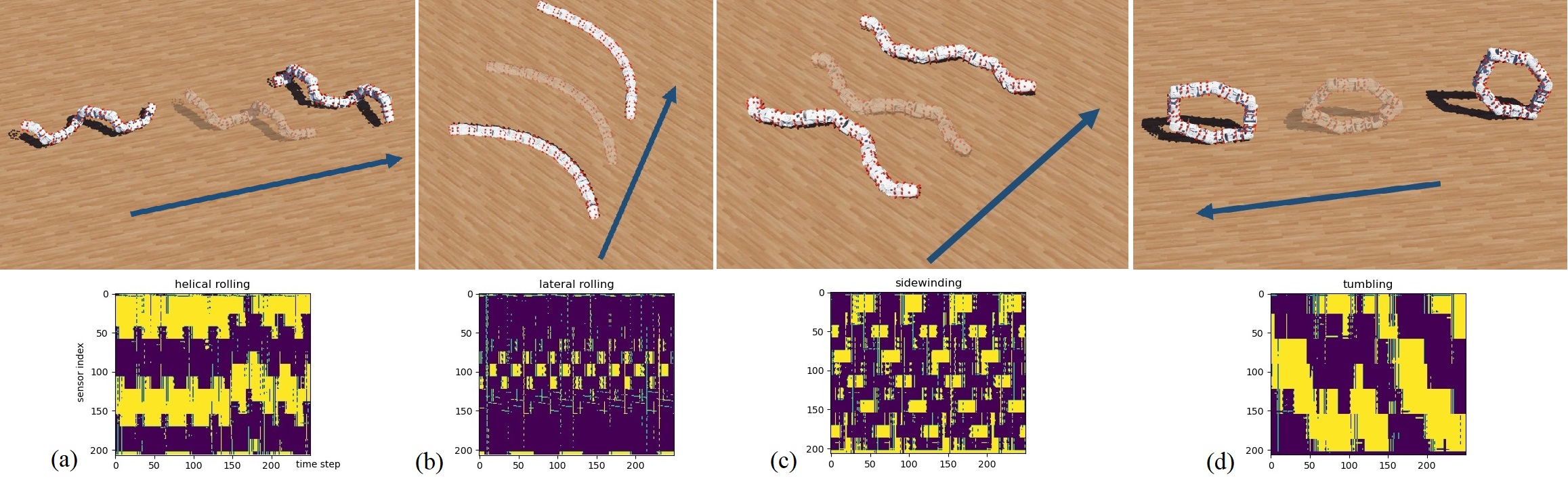}
\caption{(a) helical rolling; (b) lateral rolling; (c) sidewinding; (d) tumbling gaits and their tactile patterns.}
\label{fig_11}
\vspace{-10pt}
\end{figure*}

Deploying snake robots in more complex scenarios, such as those with obstacles or uneven terrain, makes dynamic modeling less feasible. Consequently, model-free control methods, such as reinforcement learning (RL), have garnered more extensive discussion. Additionally, adapting to complex environments relies on more comprehensive sensory patterns. Thus, some snake robots integrated various types of sensors, such as head cameras \cite{bing2020perception} and body tactile sensors \cite{liljeback2011snake, sanfilippo2016virtual, kamegawa2020three, ramesh2022snake}. Leveraging whole-body tactile sensors, snake robots can perceive the precise location and force of each contact, adjusting their body shape to avoid damage or to generate additional propulsion. Several designs for applying whole-body tactile sensing on snake robots have been proposed. However, discussions regarding the integration of tactile information into control loops are less common. Key solutions include lateral inhibition \cite{kamegawa2012proposal} or lateral hump \cite{zhen2015modeling}, relying solely on localized body curvature to adapt to contact, with less consideration for the resultant whole-body motion. When we view all surface tactile sensors as a unified entity, different gaits can produce distinct contact patterns, as shown in Fig. \ref{fig_11}. These contact patterns encapsulate substantial information about terrain and body movement, and can be used to enhance environmental perception and motion control.

In this work, we propose a hierarchical reinforcement learning (HRL) approach to control snake robots for large-scale path tracking in complex terrains, while incorporating whole-body tactile sensing into the control loop to achieve terrain adaptability. Inspired by the tactile patterns in Fig. \ref{fig_11}, we use a computer-vision-style signal processing scheme for tactile signal interpretation. Concurrently, we use curriculum learning to facilitate the expansion from small-scale solutions to large-scale scenarios, enhancing task generalization capabilities. We rely on multi-agent reinforcement learning (MARL) to provide local information exclusively to the control loop, while generating coordinated behavior under the guidance of a centralized critic. Lastly, to address the efficiency issues associated with robot simulators featuring numerous tactile sensors, we designed a distributed RL framework capable of harnessing cluster computing for acceleration.
\section{BACKGROUND}
\subsection{Markov Decision Process (MDP)}
An MDP is a 4-tuple $M=\left \langle S,A,P,R \right \rangle$ where $S$ is the set of states, $A$ is the set of actions, $P\left ( s_{t+1}| s_{t}, a_{t}\right )$ is the transition probability that action $a$ in state $s$ at time $t$ that will lead to state $s$ at time $t+1$, $R\left ( a_{t},s_{t} \right )$ is the distribution of reward when taking action $a_{t}$ in state $s_{t}$. A policy $\pi\left ( a_{t} |s_{t} \right )$ is defined as the probability distribution of choosing action $a_{t}$ given state $s_{t}$. The learning goal is to find a policy $\pi ^{*}$ that maximizes the accumulated reward in given horizon $T$, $\pi ^{*}=\underset{\pi }{\textup{argmax}}\underset{a_{t},s_{t}\sim \pi}{\mathbb{E }}\left [ \sum _{t=0}^{T-1}\gamma^{t}\cdot R\left ( a_{t},s_{t} \right ) \right ]$, where $\gamma$ is discount factor. RL algorithms are common choices to solve MDP problems.
\subsection{Central Pattern Generators (CPGs)}
CPG is a neural circuit in the vertebrate spinal cord that generates coordinated rhythmic output signals to control robot locomotion. CPG-based control methods have been successfully applied to many kinds of robots, such as multi-legged robots \cite{bellegarda2022cpg, kent2022improved, bellegarda2022visual} or snake robots \cite{liu2020learning, bing2017cpg, liu2021learning}. Usually, to improve the terrain adaptability of CPG, optimization algorithms are often applied to adjust CPG parameters in real-time. As multiple CPG structures have been proposed, we adopted the structure in \cite{bing2017cpg}. The dynamics of CPG are shown in Equation \ref{eq_1}-\ref{eq_2}.
\begin{equation}
    \begin{split}
         \dot{\varphi}=\omega +\textbf{A}\cdot \varphi +\textbf{B}\cdot \theta \\
         \ddot{r}=a\cdot \left [ \frac{a}{4} \left ( R-r \right )-\dot{r}\right ]\\
         x=r\cdot sin\left ( \varphi  \right )+\delta 
    \end{split}
    \label{eq_1}
\end{equation}
\begin{equation}
\footnotesize
    \textbf{A}=\begin{bmatrix}
 -\mu _{1}& \mu _{1} &  &  &  \\
 \mu _{2}& -2\mu _{2} & \mu _{2} &  &  \\
 &  & \ddots  &  &  \\
 &  & \mu _{n-1} & -2\mu _{n-1} & \mu _{n-1} \\
 &  &  & \mu _{n} & -\mu _{n} \\
\end{bmatrix}
\end{equation}
\begin{equation}
\footnotesize
    \textbf{B}=\begin{bmatrix}
1 &  &  &  &  \\
-1 & 1 &  &  &  \\
 & -1 & \ddots  &  &  \\
 &  &\ddots   &1  &  \\
 &  &  &-1  &  1\\
 &  &  &  &  -1\\
\end{bmatrix}
\label{eq_2}
\end{equation}
$\varphi \in \mathbb{R} ^{n}$ and $r \in \mathbb{R} ^{n}$ are internal states of CPG, $n$ is the number of output channels, typically the number of robot joints. $a$ and $\mu_{i}$ are hyperparameters that control the convergence rate. $R\in \mathbb{R} ^{n}$, $\omega\in \mathbb{R} ^{n}$, $\theta\in \mathbb{R} ^{n-1}$, $\delta\in \mathbb{R} ^{n}$ are inputs that control the desired amplitude, frequency, phase shift and offset. $x \in \mathbb{R} ^{n}$ is the output sinusoidal waves of $n$ channels.
\subsection{Robot Description}
\begin{figure}[t]
\centering
\includegraphics[width=.8\linewidth]{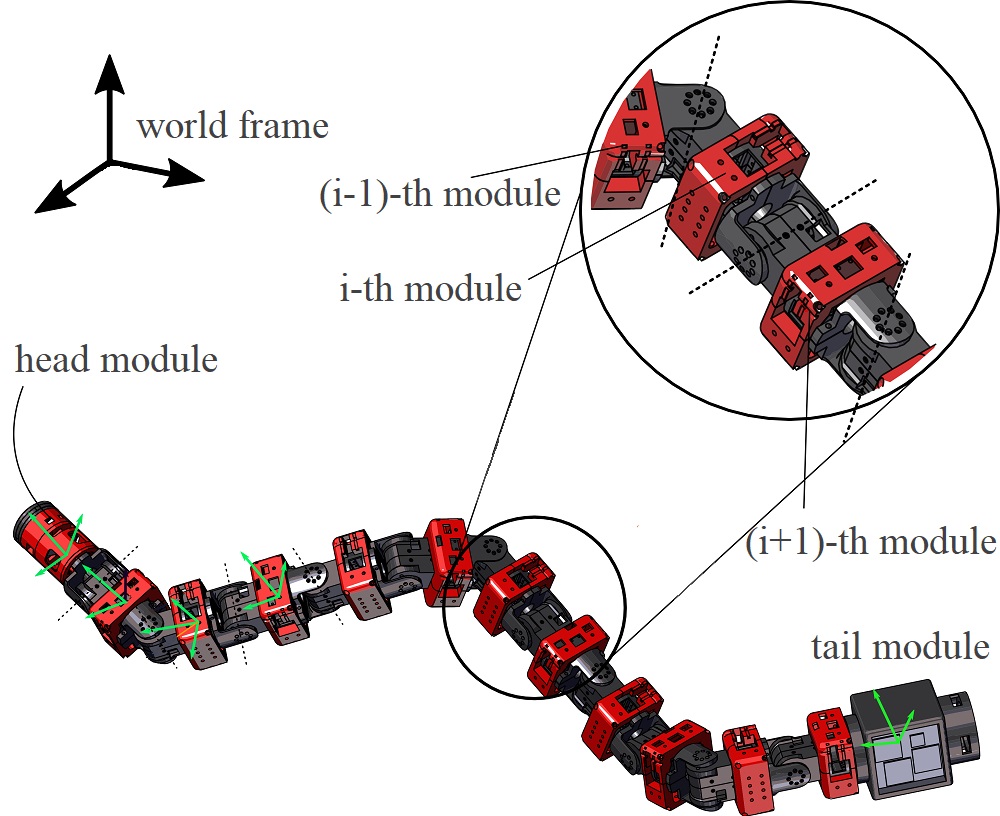}
\caption{COBRA robot body structure.}
\label{fig_20}
\end{figure}
The Crater Observing Bio-inspired Rolling Articulator (COBRA), shown in Fig.~\ref{fig_20}, is designed for space exploration \cite{wugiasoihf}. It consists of 11 actuated joints. The head module of the robot contains the onboard computing system, a radio antenna for communicating with a lunar orbiter, and an inertial measurement unit (IMU) for navigation. At the tail end, there is an interchangeable payload module containing a neutron spectrometer to detect water ice. The rest of the system consists of identical 1-DoF joint modules inter-placed perpendicularly containing a joint actuator and a battery. 207 virtual tactile sensors measuring normal pressure forces are evenly placed throughout the robot body.
\section{HIERARCHICAL CONTROL SCHEME}
\begin{figure}[t]
\centering
\includegraphics[width=.9\linewidth]{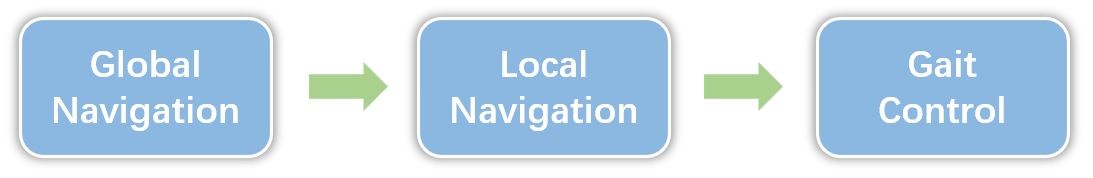}
\caption{Three-layer hierarchical control scheme.}
\label{fig_50}
\vspace{-10pt}
\end{figure}

\begin{figure}[t]
\begin{subfigure}{.45\linewidth}
    \centering
    \includegraphics[height=.10\textheight]{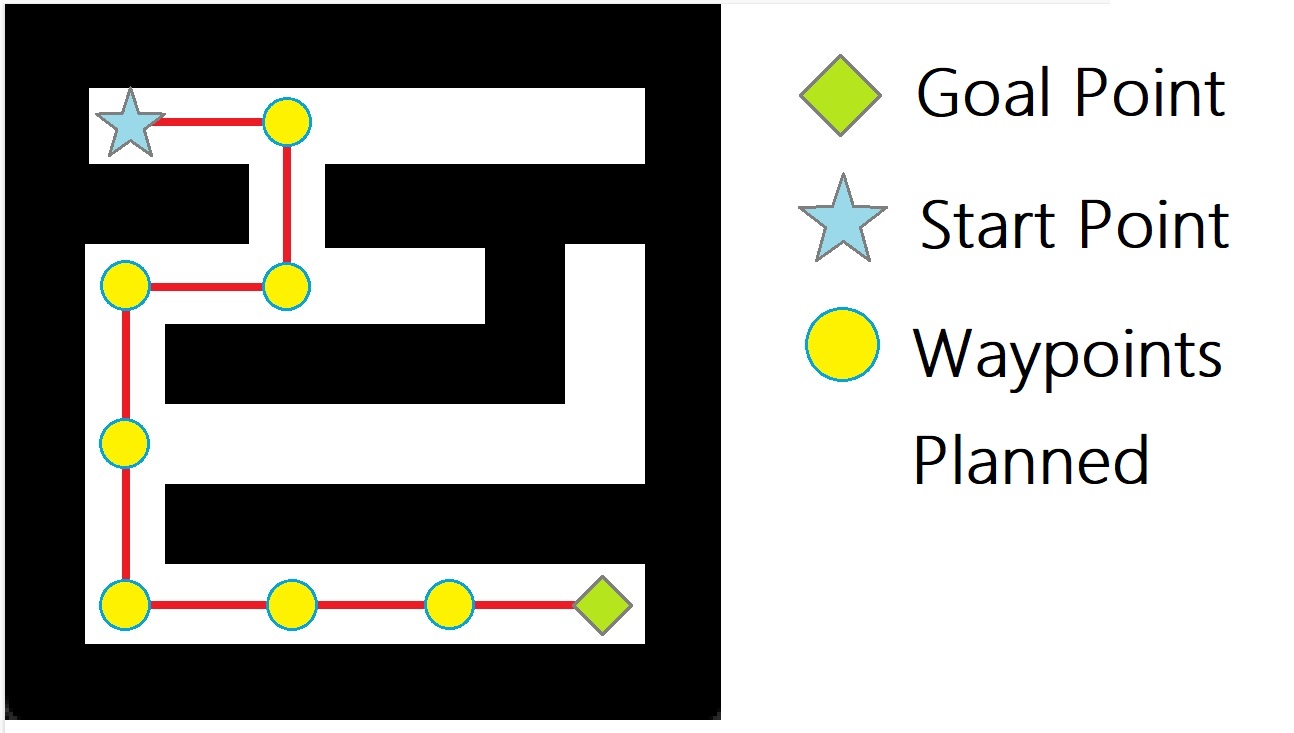}
    \caption{}
    \label{fig_39}
\end{subfigure}
\begin{subfigure}{.5\linewidth}
    \centering
    \includegraphics[height=.10\textheight]{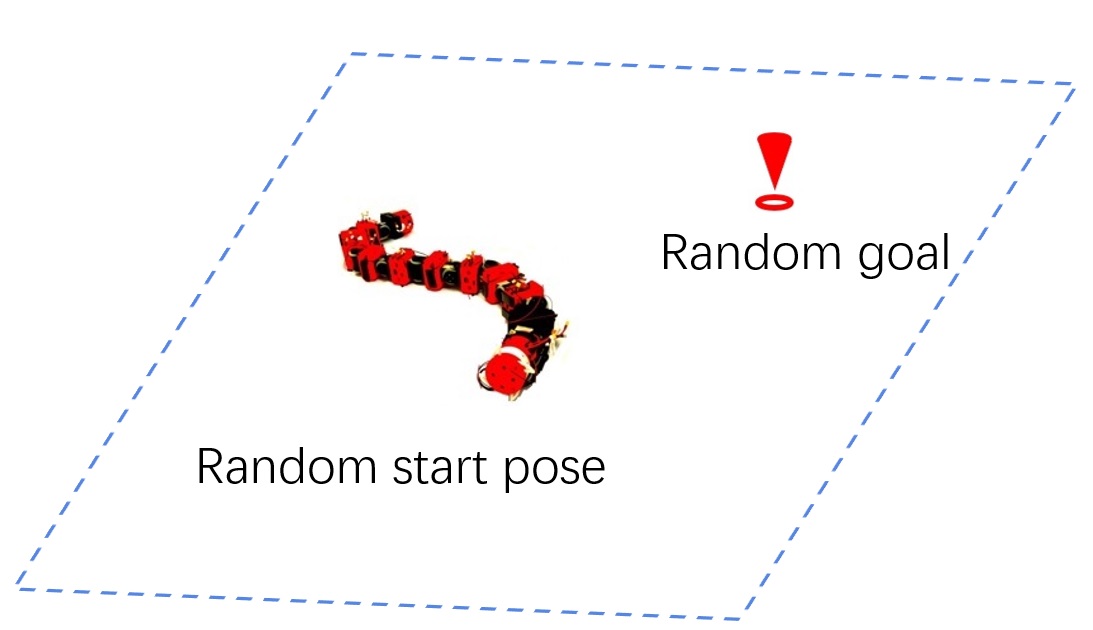}
    \caption{}
    \label{fig_38}
\end{subfigure}
\caption{(a) Global navigation: plan a path from start position to goal position and segment the path by waypoints; (b) Local navigation: using RL to steer robot from a waypoint to the next.}
\end{figure}

The problem addressed in this paper is robot navigation in complex terrains, namely, moving from any starting position to any target location on a map. To achieve this goal, we introduce a hierarchical control scheme (Fig. \ref{fig_50}). At the highest level of global navigation, we use tree search (A*) algorithm to plan efficient paths, which are then segmented into a series of contiguous waypoints (Fig. \ref{fig_39}). At the middle level, local navigation, we use  RL to train the robot to adjust its gait to navigate from one waypoint to the next (Fig. \ref{fig_38}). We integrate tactile perception information into the RL control loop to achieve real-time terrain adaptability. At the lowest level, we use  PID controllers to actuate the robot's joints to execute the desired gaits.

\begin{figure}[t]
\centering
\includegraphics[width=\linewidth]{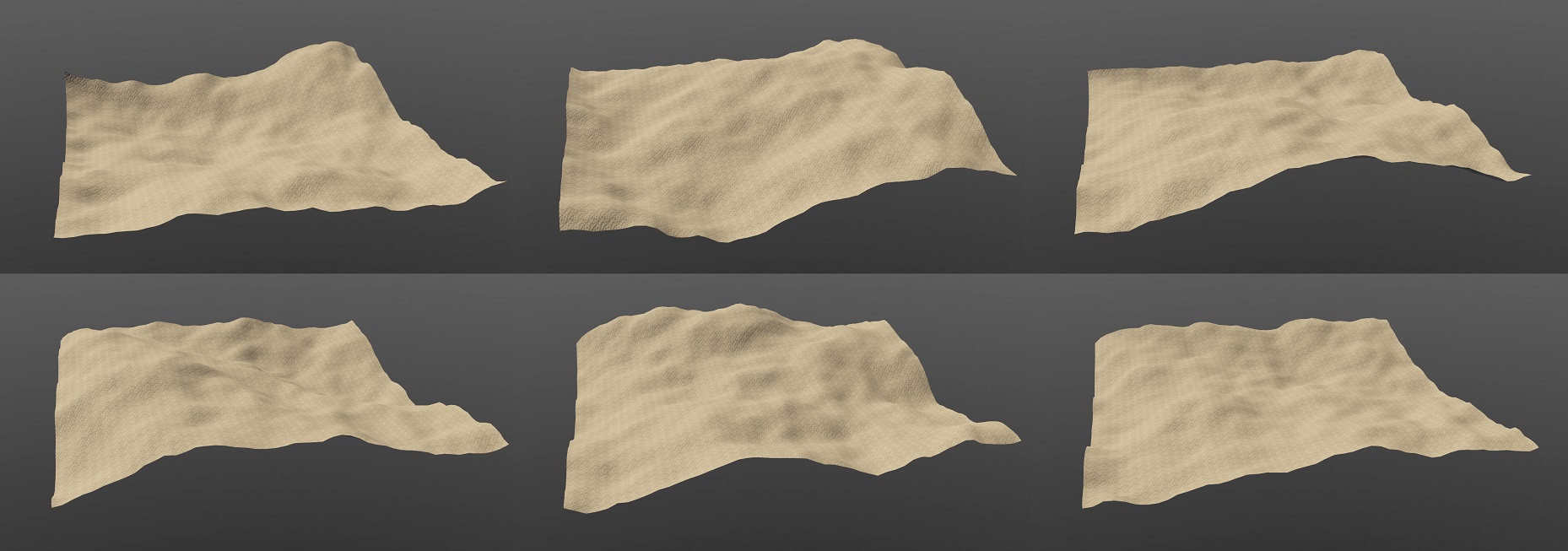}
\caption{Randomly generated curriculum terrains.}
\label{fig_temp_5}
\vspace{-10pt}
\end{figure}
\begin{figure*}[t]
\centering
\includegraphics[width=.8\linewidth]{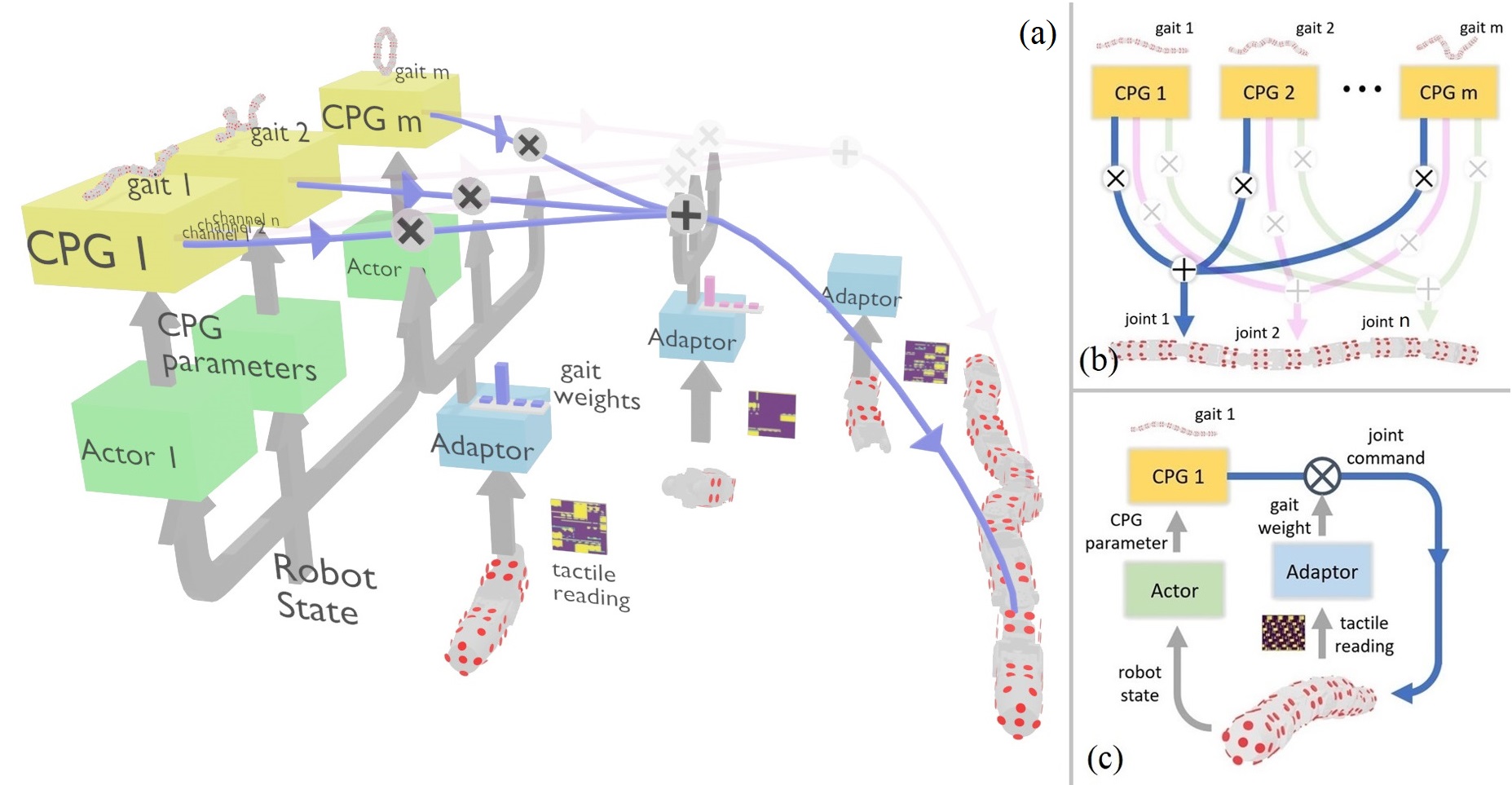}
\caption{(a) Diagram of reinforcement learning; (b) top view: each CPG module generates a particular gait pattern, consisting of $n$ channels to guides joint motions. The joint command is a linear combination of different gaits; (c) side view: each joint has its own Adaptor and the Adaptors control the gait mixing factors given the local tactile readings }
\label{fig_53}
\vspace{-10pt}
\end{figure*}
\subsection{Tactile-adaptive Local Navigation}
Given the relatively straightforward solution of global navigation and gait controller, we omit their implementation details. The novelty of this study lies in the development of a new reinforcement-learning paradigm at the local-navigation level to govern the locomotion of the snake robot. The key problem rests in effectively exploiting the whole-body tactile sensing information to regulate the robot's gait for enhanced terrain adaptability. Our control scheme design adheres to four guiding principles: (1) Individual joint control; (2) using a pre-trained gait library built from curriculum learning; (3) joint gaits solely depend on local tactile signals; (4) the application of Centralized Training and Decentralized Execution (CTDE) to mitigate partial observability and improve learning efficiency.
\subsection{Curriculum Learning}
Snake robots adopt distinct gaits for efficient locomotion on various terrains. For instance, sidewinding is often used on slopes, while lateral rolling is preferable on smoother surfaces. Inspired by this, we train agents across a spectrum of randomly generated distinctive curriculum terrains (Fig. \ref{fig_temp_5}). Each agent contains a Central Pattern Generator (CPG) module, with the actor's outputs tuning the parameters of the corresponding CPG module. By CPG parameter adjustment, the agents generate optimal gaits pertinent to their curriculum terrains. The training terrains are generated by Perlin noise of size $ 16m\times 16m$, where the robot learns to navigate to reach a random goal pose from any start location in each episode. The corresponding MDP is defined as:

\textbf{State space}: The state space includes the robot state part and tactile readings part. The robot state part consists of the joint positions $\mathbb{R}^{n}$, IMU readings $\mathbb{R}^{3}$, spatial translation between robot frame and goal pose frame $\mathbb{R}^{3}$ and relative rotation parameterized by axis-angle system $\mathbb{R}^{4}$, i.e., 21 dimensions in total. We only use ego-centric observations from the robot, so a motion capture system is not required as in \cite{transeth20083, liljeback2011experimental, hasanzadeh2010ground}, which makes our system more practical in outdoor environments.

\textbf{Action space}: The action space outputs the CPG parameters, including the desired amplitude $R\in \mathbb{R}^{n}$, frequency $\omega\in \mathbb{R}^{n}$, phase shift $\theta\in \mathbb{R}^{n-1}$ and offset $\delta\in \mathbb{R}^{n}$. 

\textbf{Reward}: We encourage the robot to reach the goal as soon as possible. The reward consists of the following terms:
\begin{equation}
    \begin{split}
        r_{1} &= \frac{1}{0.1+d_{t}}\\
        r_{2} &= d_{t-1}-d_{t}
    \end{split}
\end{equation}
\noindent where $d_{t}$ is the distance between the robot frame and the waypoint frame. $r_{1}$ encourages getting closer to the goal and $r_{2}$ encourages higher velocities. $r_{1}$ and $r_{2}$ work in a complementary fashion, with $r_{1}\rightarrow 0$ when the robot is far away from the goal and $r_{2}\rightarrow 0$ when the robot is near the goal. We use Soft Actor Critic \cite{haarnoja2018soft} as the backbone RL algorithm.

\subsection{Local Navigation Control Scheme}

Through curriculum learning as discussed earlier, we obtain a set of agents adapted to various types of terrains, achieving specific gaits by modulating the parameters of the CPG modules. The actors of all acquired agents constitute a gait library, as illustrated in the left side of Fig. \ref{fig_53}(a), represented by the yellow and green boxes. Importantly, it should be noted that the training process of these gait libraries do not involve tactile information. Our experimental findings reveal that incorporating tactile information simply by adding it to the state space of a single agent does not yield effective terrain-adaptive gaits. Hence, we devised an approach to incorporate the tactile information as in Fig. \ref{fig_53}(a), during a second phase of training (after the first phase of curriculum learning).

For each joint, we introduce an Adaptor, which takes as input the localized tactile information from adjacent links on the body, recognizes terrain features, and subsequently selects a gait output from the library in a one-hot manner. We use SAC with discrete action space to train the Adaptors, keeping the weights and biases of the Actors fixed during this training phase. We train on multiple new terrains not in the curriculum to improve the robot's terrain adaptation capabilities. In this second phase, the state space is the recent tactile readings gathered from the past one second, and the action space is the one-hot gait selection signal, and the reward is unchanged. Since the basic gaits were already learned during the curriculum learning phase, there was no need to learn gaits from scratch in this phase.

Each CPG module outputs a target joint value $\mathbf{q}_{i}\in \mathbb{R}^{n}$, $i\in \left\{ 1, \cdots m\right\}$, where $m$ is the number of CPGs and $n$ is the number of joints (channels). For each joint $j\in \left\{ 1,\cdots , n\right\}$, its target joint value shall be chosen as the $j$-th channel from one of the candidate joint values from $m$ CPG outputs. This choice is determined by the Adaptors and becomes the final target joint value to execute (Fig. \ref{fig_53}(b)).

This formulation of localized Adaptors relies on the assumption that gait adjustments are locally dependent on tactile signals, with limited reliance on distant tactile signals. For instance, the motion of a robot's head exhibits negligible correlation with the tactile feedback at its tail. Such framework draws inspiration from the Centralized Training and Decentralized Execution (CTDE) learning paradigm \cite{lyu2021contrasting, jiang2021multi} within the context of multi-agent reinforcement learning (MARL). In this analogy, akin to our Adaptors, each agent exclusively bases its decision-making process on a subset of the global observation. This configuration eliminates the redundant inter-dependencies among agents and reduces model dimensions without degrading task performance.

An intriguing observation is that when Adaptors use a softmax-based output instead of a one-hot one, the weighted mixture of gaits from the library as the final gait did not yield effective performance. The Adaptors will converge toward the average of all gaits in the library, completely neglecting the tactile information. Introducing entropy as an additional loss term could circumvent this averaging tendency but simultaneously introduces computational instability. Hence we used SAC with discrete action space to output a hard-max (one-hot) gait selection.

\begin{table}[t]
    \centering
     \begin{tabular}{|m{4.0em}|m{2.1em}|m{2.1em}|m{2.1em}|m{2.1em}|m{2.1em}|} 
     \arrayrulecolor{gray}
     \hline
     \rowcolor{LightCyan}
     Num. sensors & 0 & 50 & 100 & 150 & 200  \\ 
     \hline\hline
     \rowcolor{light-gray}
      \cellcolor[rgb]{0.772549, 0.878431, 0.705882}Gazebo& 
      2.28 & 
      0.27 & 
      0.09 & 
      0.05 & 
      0.02\\
     \rowcolor{light-gray}
     \cellcolor[rgb]{0.772549, 0.878431, 0.705882}Mujoco& 
      110.3 & 
      31.79 & 
      24.37 & 
      19.92 & 
      12.37\\
     \rowcolor{light-gray}
     \cellcolor[rgb]{0.772549, 0.878431, 0.705882}Webots& 
      42.3 & 
      2.31 & 
      1.08 & 
      0.69 & 
      0.33\\
     \rowcolor{light-gray}
     \cellcolor[rgb]{0.772549, 0.878431, 0.705882}PyBullet& 
      59.4 & 
      49.8 & 
      34.8 & 
      NaN & 
      NaN\\
     \hline
    \end{tabular}
    \caption{Real Time Factor (RTF) comparison among popular simulators. NaN represents unstable calculation.}
    \label{tab_3}
    \vspace{-10pt}
\end{table}

\begin{figure}[t]
\centering
\includegraphics[width=.8\linewidth]{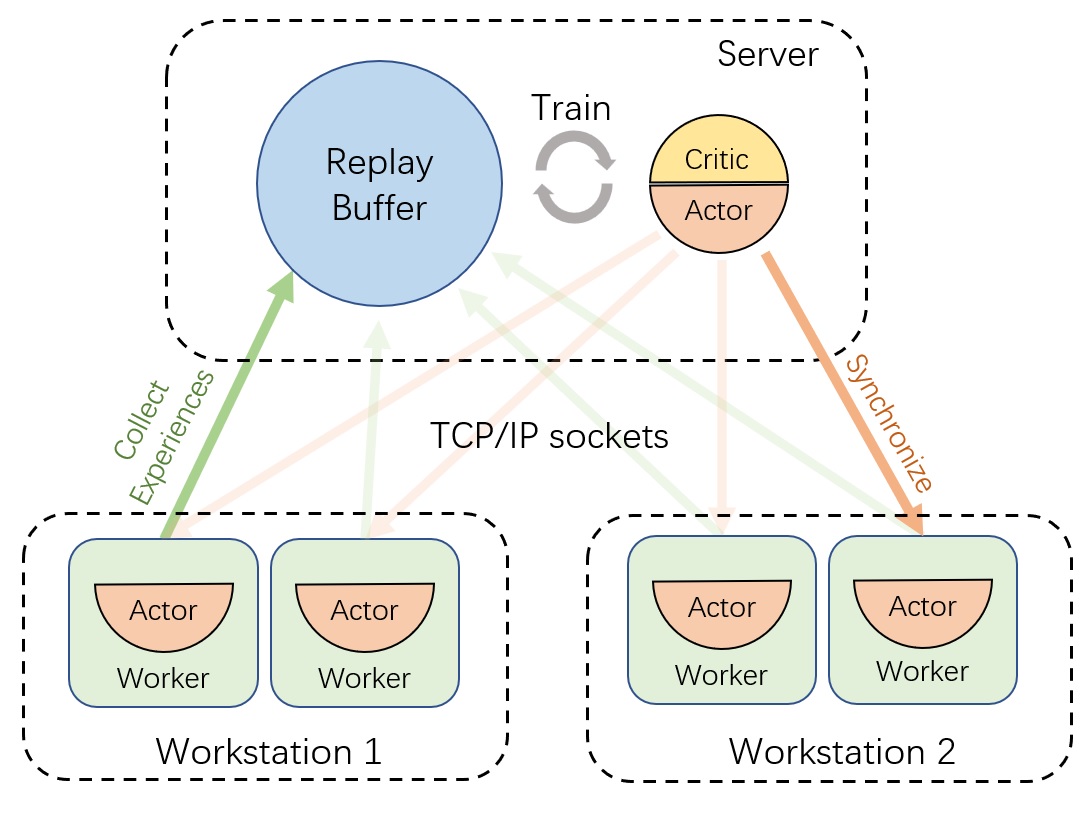}
\caption{Distributed reinforcement learning framework.}
\label{fig_7}
\vspace{-10pt}
\end{figure}

\begin{figure}[t]
\centering
\includegraphics[width=\linewidth]{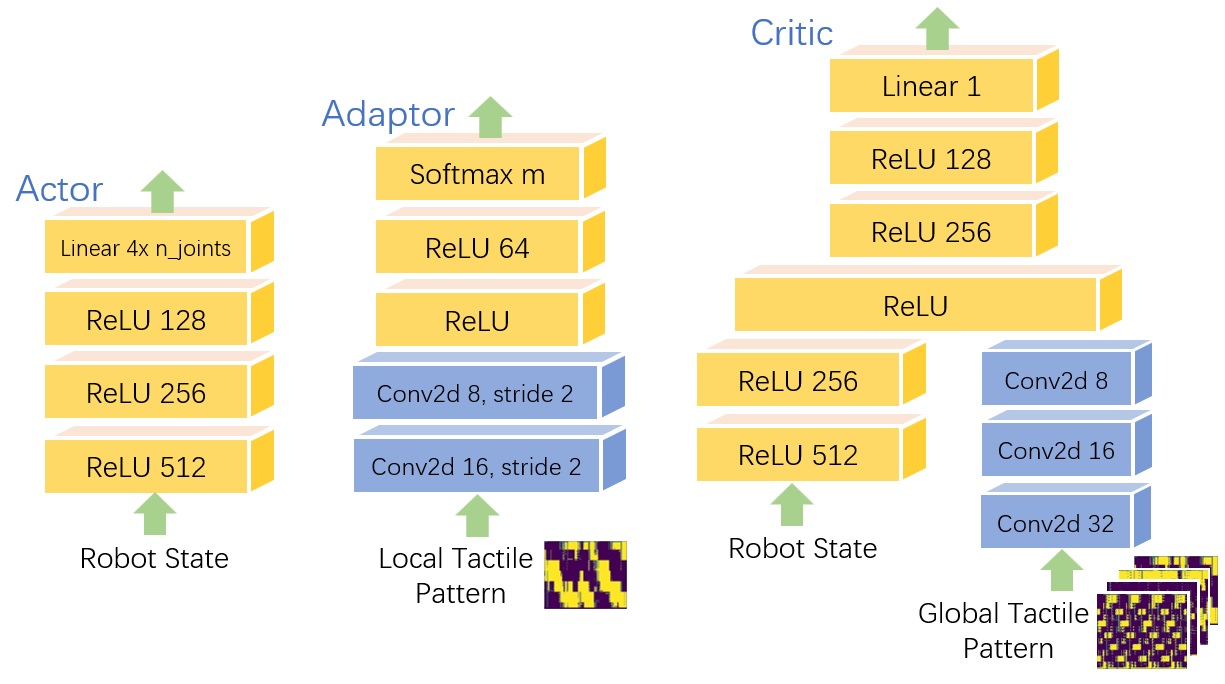}
\caption{Neural network architectures.}
\label{fig_44}
\vspace{-10pt}
\end{figure}

\begin{figure}[t]
\centering
\includegraphics[width=\linewidth]{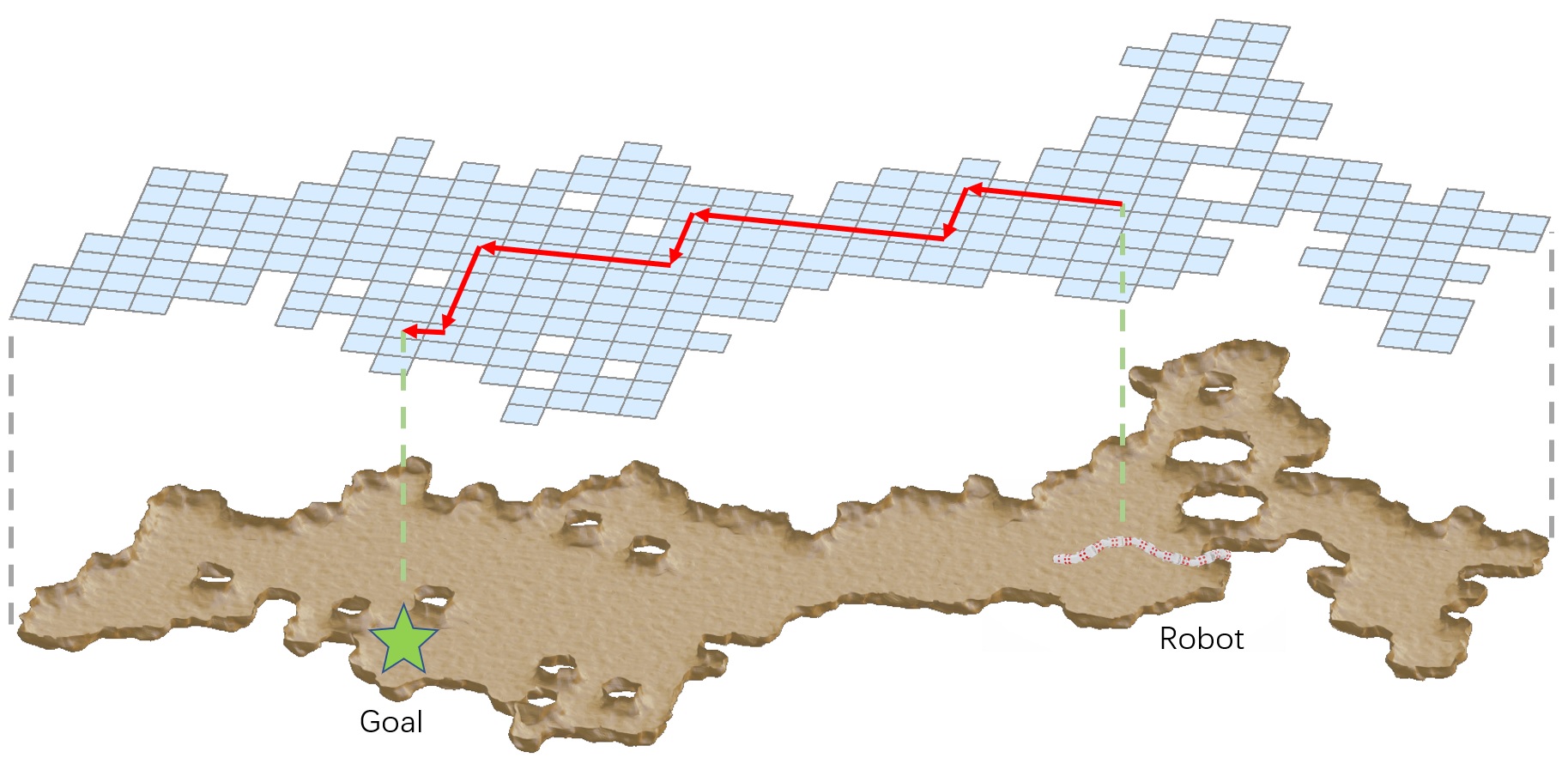}
\caption{Randomly generated cave as the test domain. Global navigation plans a path from current robot location to the goal (red arrows), and local navigation controls robot gaits to traverse between neighborhood blocks.}
\label{fig_33}
\vspace{-10pt}
\end{figure}

\begin{figure}[t]
\centering
\includegraphics[width=\linewidth]{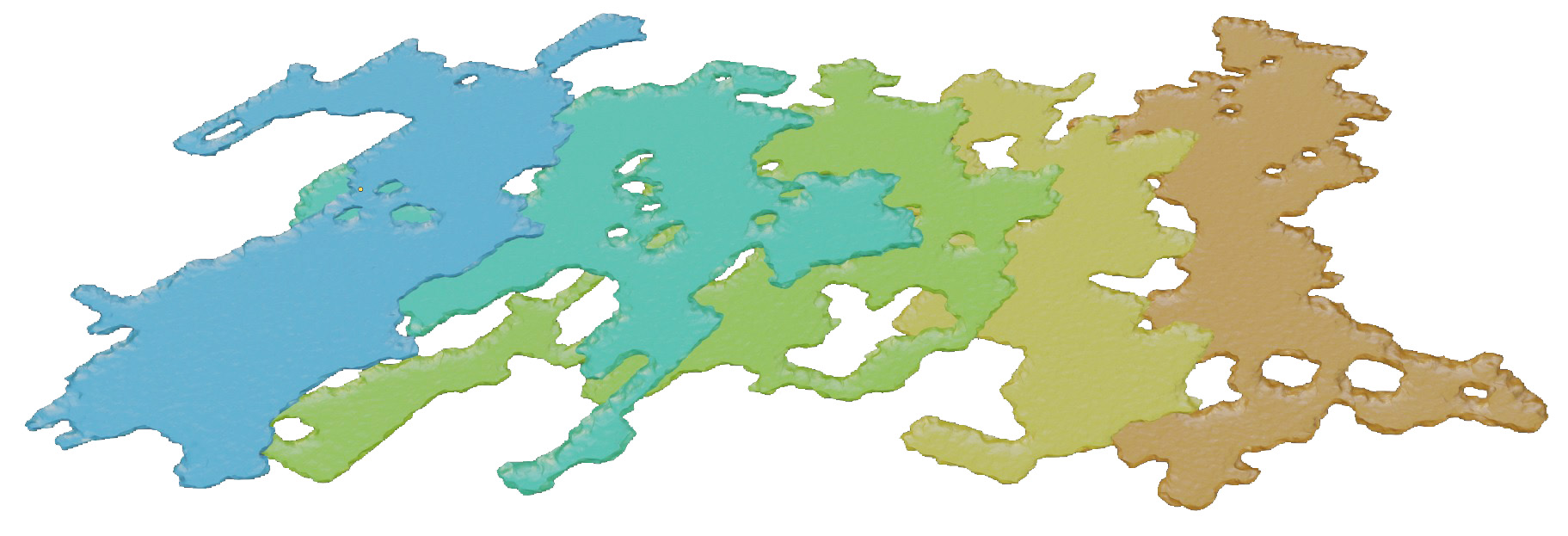}
\caption{Randomly generated cave layouts.}
\label{fig_45}
\vspace{-10pt}
\end{figure}

\subsection{Distributed Learning}

Due to the introduction of tactile sensors, simulation becomes slow and does not scale well to the substantial amount of experience 
required in RL. Table \ref{tab_3} illustrates the operational efficiency of several commonly used robot simulators concerning various numbers of tactile sensors. It can be observed that as the number of sensors increases, there is a noticeable decline in the simulator's efficiency, as manifested by the maximum real-time acceleration achievable by the simulator, denoted as the Real-Time Factor (RTF). Therefore, we developed a distributed RL framework deployable across multiple workstations (Fig. \ref{fig_7}) to mitigate this situation. One of these workstations serves as a server, with an agent comprising a critic and an actor (gait library and Adaptors), along with a centralized replay buffer to store experiences. The other workstations run multiple simulator instances (workers), each instance containing only one agent interacting with the environment. The experiences gained by the workers are transmitted to the server via the TCP/IP protocol, and agent training is exclusively conducted at the server end. The server periodically synchronizes the actors to each worker. Notably, each Adaptor only receives a local tactile pattern, recorded from two adjacent links of a joint, while the critic receives global tactile patterns from all sensors across the entire body. The neural networks architectures we use are shown in Fig. \ref{fig_44}.

\section{EXPERIMENTS}

We tested the terrain adaptability of snake robot locomotion in a randomly generated cave, as shown in Fig. \ref{fig_33}. The dimensions of the cave are 155m$\times$102m. The uneven surface of the cave presents challenges for the robot to move. The task involves autonomous navigation of the robot from any initial position to any specified target point, using terrain-adaptive locomotion. We divided the cave into 4m$\times$4m blocks, and the high level controller plans paths based on this grid (Fig.~\ref{fig_33}). To show the generalization ability, we generate multiple random cave layouts as shown in Fig. \ref{fig_45}.

\begin{figure}[t]
\centering
\includegraphics[width=\linewidth]{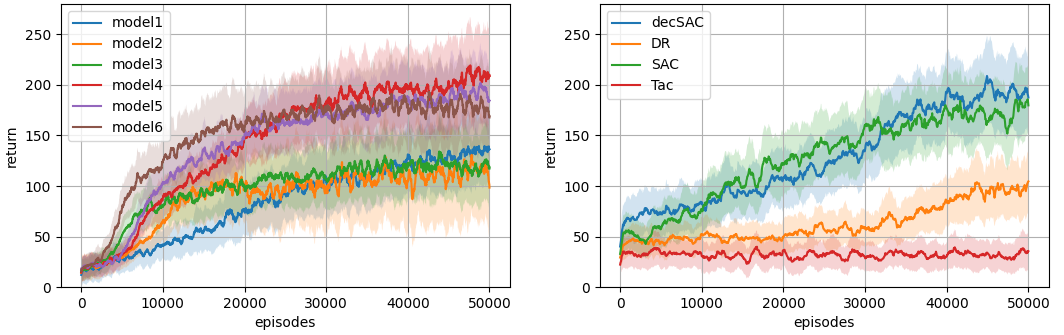}
\caption{Training curves: (left) curriculum training phase; (right) tactile-adaptation phase.}
\label{fig_49}
\end{figure}

\subsection{Curriculum Results}

The training curves for the two phases of our designed algorithms are shown in Fig. \ref{fig_49}. On the left are the results of 6 curriculum learning on different terrains, during which the robot learns basic gaits without the use of tactile perception. On the right are the results of our terrain adaptation method (decSAC) trained on 6 terrains beyond the curriculum learning. It can be observed that at the beginning of the second phase, due to change in terrains, the gaits learned in the first phase are not readily adaptable to the new environment. However, after training, our algorithm demonstrated performance similar to curriculum learning on various new terrains. Additionally, we observed that there is little difference in the final performance between centralized and decentralized Adaptors (SAC vs decSAC), thus demonstrating the feasibility of training using MARL. For the method that does not use tactile information but relies solely on domain randomization (DR), it can be observed that learning is possible to some extent, but there are performance bottlenecks. Furthermore, we can see that directly incorporating tactile information as part of the state space (Tac) yields ineffective results. All the results are averaged from 10 independent trials.

Analysis of terrain adaptability can be referenced in TABLE \ref{tab_1}, where M1-M6 represent the 6 models in the curriculum training, and T1-T6 correspond to the matching training terrains for M1-M6. As observed, the diagonal shape in the table indicates that M1-M6 only perform well in their respective training scenarios but are hard to adapt to untrained environments. T7 and T8 are two entirely new test environments beyond the two training phases. It can be seen that neither M1-M6 nor DR can perform well in the new environments, whereas our approach is capable of extracting terrain characteristics from tactile information and adopting adaptive gaits.
\subsection{Cave Navigation Performance}
\begin{figure}[t]
\centering
\includegraphics[width=\linewidth]{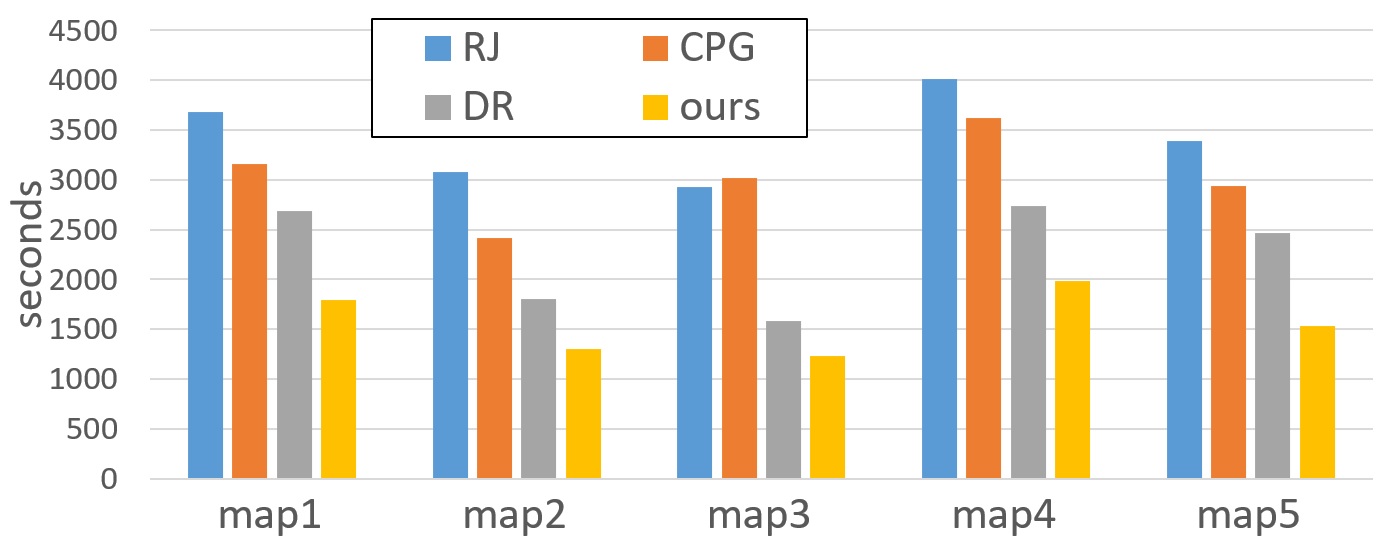}
\caption{Baseline Comparisons in the 5 test caves}
\label{fig_54}
\vspace{-10pt}
\end{figure}

We compared the results of several baselines in navigating through the five caves (Fig. \ref{fig_45}). The comparison of their runtime is shown in Fig.\ref{fig_54}. The action space of method "RJ" is the robot's target joint angles \cite{bing2020perception}, while the action space of method "CPG" consists of parameters for the CPG modules. The "DR" method introduces Domain Randomization on top of CPG. The baselines in the figure do not utilize tactile information. It can be observed that our method achieved the most efficient navigation results. We found that similar to the Tac results in Fig. \ref{fig_49}, directly incorporating tactile information into the state space, regardless of using RJ, CPG, or DR in the action space, failed to complete the navigation task within a reasonable timeframe, and therefore, the results are not depicted in the figure. We speculate that the reason might be the inherent difficulty of simultaneously learning both gait and terrain adaptability from scratch. In contrast, our approach, through curriculum learning, divides the training into two phases, each focusing on learning gait and terrain adaptability, respectively. This approach simplifies the problem by decoupling the two tasks.

\begin{table}[t]
    \centering
    \scriptsize
    \arrayrulecolor{gray}
     \begin{tabular}{|m{1.3em}|m{2.1em}|m{2.1em}|m{2.1em}|m{2.1em}|m{2.1em}|m{2.1em}|m{2.1em}|m{2.1em}|} 
     \hline
     & T1 & T2 & T3 & T4 & T5 & T6& T7 & T8  \\ 
     \hline
      \makecell[tl]{M1}& 
      \cellcolor[rgb]{0.0313, 0.2235, 0.4196}\makecell[tl]{\color{light-gray}{227.8}\\ \color{light-gray}{$\pm$ 6.1}} & 
      \cellcolor[rgb]{0.8274, 0.8901, 0.9529}\makecell[tl]{75.3\\ $\pm$ 7.3} & 
      \cellcolor[rgb]{0.8705, 0.9215, 0.9686}\makecell[tl]{59.5\\ $\pm$ 5.6} & 
      \cellcolor[rgb]{0.6901, 0.8196, 0.9058}\makecell[tl]{90.6\\ $\pm$ 9.0} & 
      \cellcolor[rgb]{0.6588, 0.8078, 0.8941}\makecell[tl]{94.2\\ $\pm$ 5.8} &
      \cellcolor[rgb]{0.7921, 0.8666, 0.9411}\makecell[tl]{76.4\\ $\pm$ 4.6} & 
      \cellcolor[rgb]{0.6745, 0.8156, 0.9019}\makecell[tl]{103.2\\ $\pm$ 4.6}& 
      \cellcolor[rgb]{0.7725, 0.8549, 0.9333}\makecell[tl]{88.1\\ $\pm$ 4.6}\\
     \hline
     \makecell[tl]{M2}&  
     \cellcolor[rgb]{0.3686, 0.6470, 0.8196}\makecell[tl]{124.9\\ $\pm$ 7.6} &
     \cellcolor[rgb]{0.0470, 0.3372, 0.6274}\makecell[tl]{\color{light-gray}{206.3}\\ \color{light-gray}{$\pm$ 3.8}} & 
     \cellcolor[rgb]{0.7803, 0.8588, 0.9372}\makecell[tl]{78.5\\ $\pm$ 6.7} & 
     \cellcolor[rgb]{0.7333, 0.8392, 0.9215}\makecell[tl]{84.5\\ $\pm$ 4.0} & 
     \cellcolor[rgb]{0.5960, 0.7803, 0.8745}\makecell[tl]{101.2\\ $\pm$ 6.4} &
     \cellcolor[rgb]{0.7490, 0.8470, 0.9254}\makecell[tl]{82.8\\ $\pm$ 8.8}&
     \cellcolor[rgb]{0.8745, 0.9215, 0.9686}\makecell[tl]{62.5\\ $\pm$ 6.4}&
     \cellcolor[rgb]{0.8156, 0.8823, 0.9490}\makecell[tl]{77.8\\ $\pm$ 6.4}\\
     \hline
      \makecell[tl]{M3}&
      \cellcolor[rgb]{0.2509, 0.5647, 0.7725}\makecell[tl]{139.4\\ $\pm$ 4.1} &
      \cellcolor[rgb]{0.8470, 0.9058, 0.9607}\makecell[tl]{64.3\\ $\pm$ 5.5} & 
      \cellcolor[rgb]{0.1019, 0.4117, 0.6823}\makecell[tl]{\color{light-gray}{163.0}\\ \color{light-gray}{$\pm$ 5.4}} & 
      \cellcolor[rgb]{0.5764, 0.7686, 0.8705}\makecell[tl]{103.2\\ $\pm$ 7.2} & 
      \cellcolor[rgb]{0.6392, 0.8, 0.8901}\makecell[tl]{96.3\\ $\pm$ 6.7} &
      \cellcolor[rgb]{0.8470, 0.9058, 0.9607}\makecell[tl]{70.4\\ $\pm$ 4.4}&
      \cellcolor[rgb]{0.7568, 0.8509, 0.9294}\makecell[tl]{90.3\\ $\pm$ 4.4}&
      \cellcolor[rgb]{0.7960, 0.8705, 0.9411}\makecell[tl]{82.8\\ $\pm$ 4.4}\\
     \hline
      \makecell[tl]{M4}&
      \cellcolor[rgb]{0.5215, 0.7372, 0.8588}\makecell[tl]{108.6\\ $\pm$ 5.7} & 
      \cellcolor[rgb]{0.6196, 0.7921, 0.8823}\makecell[tl]{98.7\\ $\pm$ 8.4} &
      \cellcolor[rgb]{0.8705, 0.9215, 0.9686}\makecell[tl]{59.8\\ $\pm$ 7.7} & 
      \cellcolor[rgb]{0.1607, 0.4745, 0.7254}\makecell[tl]{\color{light-gray}{153.0}\\ \color{light-gray}{$\pm$ 4.0}} &
      \cellcolor[rgb]{0.7529, 0.8470, 0.9294}\makecell[tl]{82.4\\ $\pm$ 6.4} &
      \cellcolor[rgb]{0.6392, 0.8, 0.8901}\makecell[tl]{96.4\\ $\pm$ 5.5}&
      \cellcolor[rgb]{0.7960, 0.8705, 0.9411}\makecell[tl]{83.1\\ $\pm$ 5.5}&
      \cellcolor[rgb]{0.6901, 0.8196, 0.9058}\makecell[tl]{101.0\\ $\pm$ 5.5}\\
     \hline
      \makecell[tl]{M5}&
      \cellcolor[rgb]{0.8078, 0.8784, 0.9450}\makecell[tl]{72.8\\ $\pm$ 6.3} & 
      \cellcolor[rgb]{0.7882, 0.8666, 0.9411}\makecell[tl]{76.5\\ $\pm$ 4.5} & 
      \cellcolor[rgb]{0.9686, 0.9843, 1.0}\makecell[tl]{40.1\\ $\pm$ 7.2} & 
      \cellcolor[rgb]{0.7803, 0.8588, 0.9372}\makecell[tl]{78.4\\ $\pm$ 4.9}& 
      \cellcolor[rgb]{0.1215, 0.4352, 0.7019}\makecell[tl]{\color{light-gray}{159.1}\\ \color{light-gray}{$\pm$ 7.0}} &
      \cellcolor[rgb]{0.7294, 0.8392, 0.9176}\makecell[tl]{85.1\\ $\pm$ 8.5}&
      \cellcolor[rgb]{0.9529, 0.9725, 0.9921}\makecell[tl]{43.8\\ $\pm$ 8.5}&
      \cellcolor[rgb]{0.8980, 0.9372, 0.9764}\makecell[tl]{56.6\\ $\pm$ 8.5}\\
     \hline
    \makecell[tl]{M6}&
     \cellcolor[rgb]{0.6392, 0.8, 0.8901}\makecell[tl]{96.3\\ $\pm$ 4.3}&
     \cellcolor[rgb]{0.7450, 0.8431, 0.9254}\makecell[tl]{83.6\\ $\pm$ 6.5}&
     \cellcolor[rgb]{0.8313, 0.8941, 0.9529}\makecell[tl]{67.4\\ $\pm$ 7.0}&
     \cellcolor[rgb]{0.6392, 0.8, 0.8901}\makecell[tl]{96.7\\ $\pm$ 6.2}&
     \cellcolor[rgb]{0.8156, 0.8823, 0.9490}\makecell[tl]{71.4\\ $\pm$ 5.9}&
     \cellcolor[rgb]{0.3058, 0.6039, 0.7921}\makecell[tl]{\color{light-gray}{154.8}\\ \color{light-gray}{$\pm$ 4.1}}&
     \cellcolor[rgb]{0.8352, 0.8980, 0.9568}\makecell[tl]{72.3\\ $\pm$ 5.9}&
     \cellcolor[rgb]{0.7686, 0.8549, 0.9333}\makecell[tl]{89.2\\ $\pm$ 5.9}\\
     \hline
     \makecell[tl]{DR}&
     \cellcolor[rgb]{0.5764, 0.7686, 0.8705}\makecell[tl]{116.3\\ $\pm$ 7.2}&
     \cellcolor[rgb]{0.6745, 0.8156, 0.9019}\makecell[tl]{102.9\\ $\pm$ 10.3}&
     \cellcolor[rgb]{0.8313, 0.8941, 0.9529}\makecell[tl]{73.4\\ $\pm$ 6.4}&
     \cellcolor[rgb]{0.5450, 0.7529, 0.8666}\makecell[tl]{120.2\\ $\pm$ 8.2}&
     \cellcolor[rgb]{0.6549, 0.8078, 0.8941}\makecell[tl]{107.5\\ $\pm$ 4.6}&
     \cellcolor[rgb]{0.3843, 0.6588, 0.8235}\makecell[tl]{140.3\\ $\pm$ 5.3}&
     \cellcolor[rgb]{0.7098, 0.8274, 0.9137}\makecell[tl]{98.9\\ $\pm$ 7.0}&
     \cellcolor[rgb]{0.2784, 0.5843, 0.7843}\makecell[tl]{\color{light-gray}{158.2}\\ \color{light-gray}{$\pm$ 7.1}}\\
     \hline
     \makecell[tl]{Ours}&
     \cellcolor[rgb]{0.0313, 0.3176, 0.6117}\makecell[tl]{\color{light-gray}{210.5}\\ \color{light-gray}{$\pm$ 13.3}}&
     \cellcolor[rgb]{0.0313, 0.1882, 0.4196}\makecell[tl]{\color{light-gray}{230.8}\\ \color{light-gray}{$\pm$ 8.7}}&
     \cellcolor[rgb]{0.6941, 0.8235, 0.9058}\makecell[tl]{101.6\\ $\pm$ 8.5}&
     \cellcolor[rgb]{0.1843, 0.4980, 0.7372}\makecell[tl]{\color{light-gray}{172.6}\\ \color{light-gray}{$\pm$ 6.7}}&
     \cellcolor[rgb]{0.1960, 0.5098, 0.7450}\makecell[tl]{\color{light-gray}{169.8}\\ \color{light-gray}{$\pm$ 9.9}}&
     \cellcolor[rgb]{0.3058, 0.6039, 0.7921}\makecell[tl]{\color{light-gray}{152.2}\\ \color{light-gray}{$\pm$ 4.6}}&
     \cellcolor[rgb]{0.4823, 0.7176, 0.8509}\makecell[tl]{127.4\\ $\pm$ 6.7}&
     \cellcolor[rgb]{0.0313, 0.1882, 0.4196}\makecell[tl]{\color{light-gray}{235.0}\\ \color{light-gray}{$\pm$ 8.8}}\\
     \hline
    \end{tabular}
    \caption{Model-terrain generalization analysis (return with standard deviation).}
    \label{tab_1}
    \vspace{-10pt}
\end{table}

\begin{figure}[t]
\centering
\includegraphics[width=.9\linewidth]{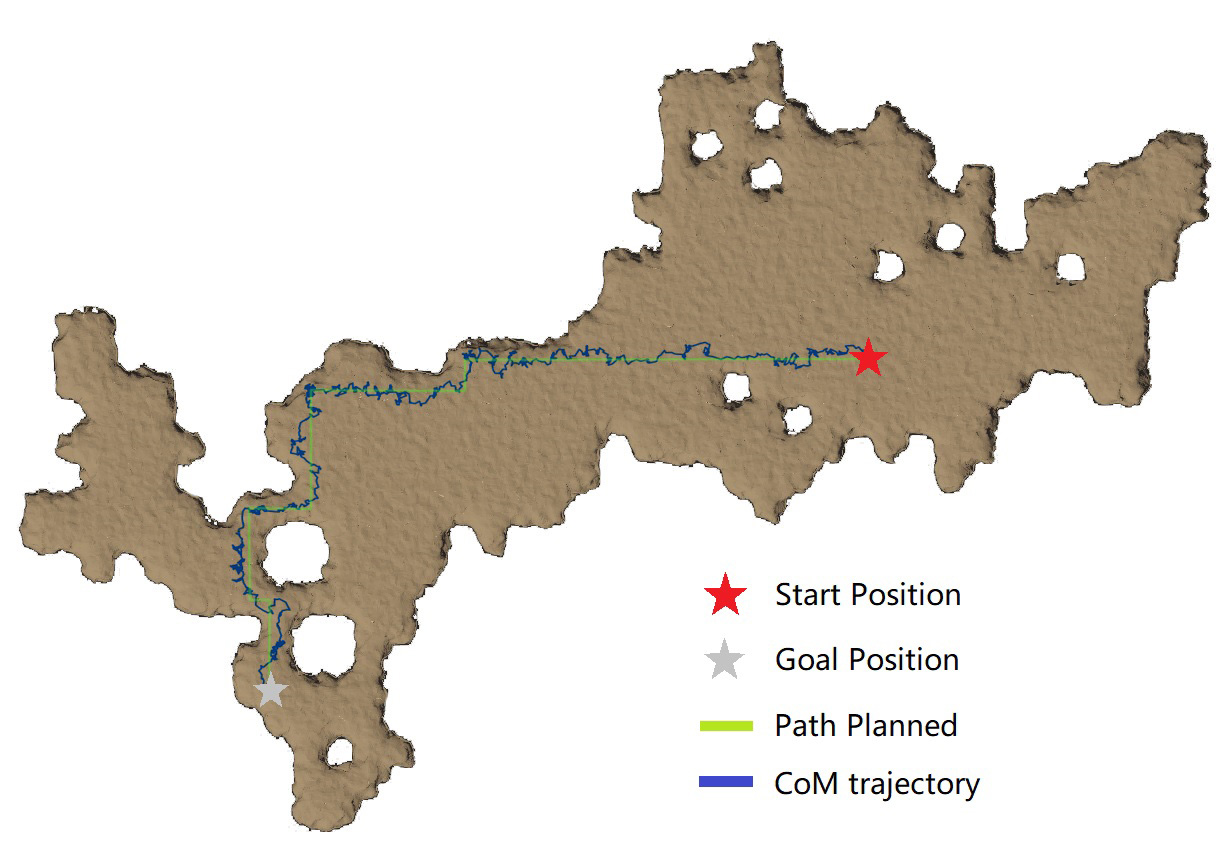}
\caption{Center of Mass (CoM) trajectory.}
\label{fig_48}
\end{figure}

\begin{figure}[h!]
\begin{subfigure}{.24\linewidth}
    \centering
    \includegraphics[height=.08\textheight]{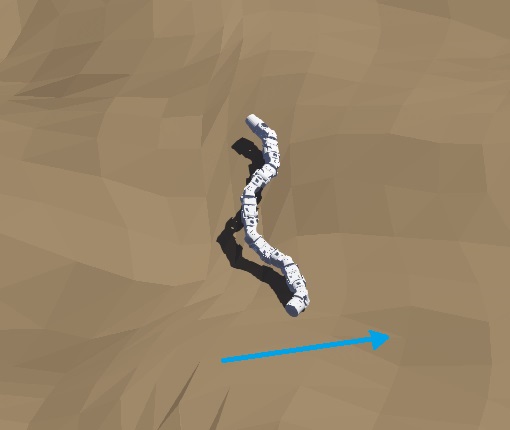}
\end{subfigure}
\begin{subfigure}{.24\linewidth}
    \centering
    \includegraphics[height=.08\textheight]{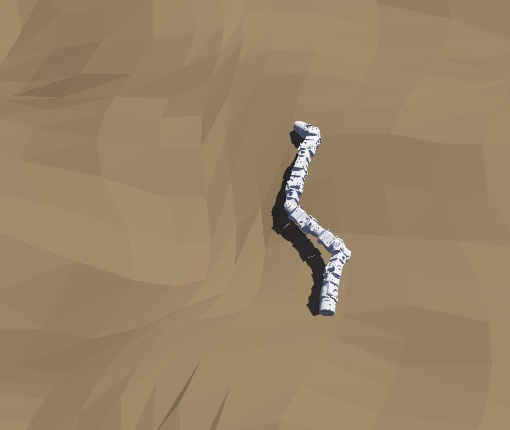}
\end{subfigure}
\begin{subfigure}{.24\linewidth}
    \centering
    \includegraphics[height=.08\textheight]{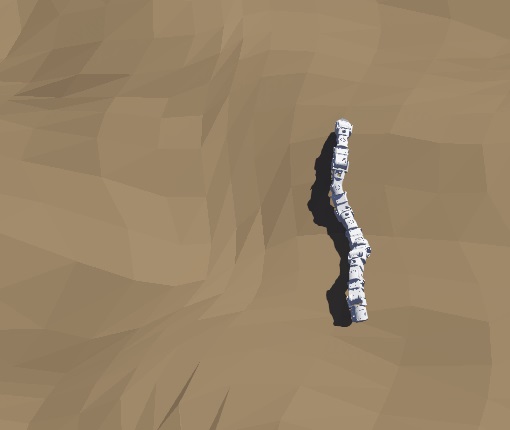}
\end{subfigure}
\begin{subfigure}{.24\linewidth}
    \centering
    \includegraphics[height=.08\textheight]{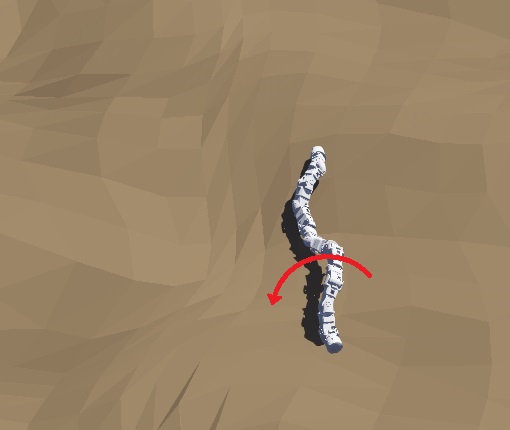}
\end{subfigure}
\caption{Tactile-less CPG controller has difficulty traversing the terrain with sidewinding gait, the robot falls from uphill climbing.}
\label{fig_43}
\vspace{-10pt}
\end{figure}

The centroid motion trajectory of the robot in one of the caves is shown in Fig. \ref{fig_48}, and it can be observed that the centroid motion trajectory closely aligns with the path planned by the high-level controller. By observing the robot's motion at a closer distance, we found that when tactile information is not utilized, i.e., RL directly determines the parameters of the CPG module based on the robot's state, the terrain adaptability is compromised. As shown in Fig. \ref{fig_43}, when the robot uses the sidewinding gait on an uphill without the slope information, it is prone to turning over. We provide a detailed demonstration of this comparison in the supplementary video, showcasing the motion of various baselines within the caves.

\section{CONCLUSION}
In this paper, we proposed a novel hierarchical reinforcement learning control scheme to address the navigation problem of snake robots equipped with whole-body tactile perception in complex terrains. By incorporating tactile information, snake robots can perceive environment characteristics and adjust their gaits accordingly to achieve terrain adaptability. Validation experiments across various terrains demonstrated superior performance of our approach compared to traditional RL solutions. Future work will focus on sim-to-real transfer and conducting validation on real robot platforms.

\newpage

\bibliographystyle{IEEEtran}
\bibliography{IEEEabrv,mybibfile,alireza-refs}

\end{document}